\documentclass[10pt,twocolumn,letterpaper]{article}

\usepackage[pagenumbers]{cvpr}              
\usepackage[accsupp]{axessibility}

\usepackage[dvipsnames]{xcolor}

\definecolor{cvprblue}{rgb}{0.21,0.49,0.74}
\usepackage[pagebackref,breaklinks,colorlinks,citecolor=cvprblue]{hyperref}
\usepackage{booktabs}
\usepackage{multirow}
\usepackage{graphicx}
\usepackage{pythonhighlight}
\usepackage{color}
\usepackage{listings}
\usepackage[normalem]{ulem}
\useunder{\uline}{\ul}{}
\def\paperID{12651} 
\def\confName{CVPR}
\def\confYear{2024}

\title{Choose What You Need: Disentangled Representation Learning for \\ Scene Text Recognition, Removal and Editing}

\author{Boqiang Zhang \quad Hongtao Xie\footnotemark[1] \quad Zuan Gao \quad Yuxin Wang\\
University of Science and Technology of China\\
{\tt\small \{cyril,zuangao\}@mail.ustc.edu.cn \quad \{htxie,wangyx58\}@ustc.edu.cn}
}

\begin{document}
\maketitle
\renewcommand{\thefootnote}{\fnsymbol{footnote}}
\footnotetext[1]{The corresponding author.}
\renewcommand{\thefootnote}{\arabic{footnote}}
\begin{abstract}
Scene text images contain not only style information (font, background) but also content information (character, texture). Different scene text tasks need different information, but previous representation learning methods use tightly coupled features for all tasks, resulting in sub-optimal performance. We propose a \underline{D}isent\underline{a}ngled \underline{R}epresentation \underline{L}earn\underline{ing} framework (DARLING) aimed at disentangling these two types of features for improved adaptability in better addressing various downstream tasks (choose what you really need).
Specifically, we synthesize a dataset of image pairs with identical style but different content.
Based on the dataset, we decouple the two types of features by the supervision design. Clearly, we directly split the visual representation into style and content features, the content features are supervised by a text recognition loss, while an alignment loss aligns the style features in the image pairs.
Then, style features are employed in reconstructing the counterpart image via an image decoder with a prompt that indicates the counterpart's content. Such an operation effectively decouples the features based on their distinctive properties. To the best of our knowledge, this is the first time in the field of scene text that disentangles the inherent properties of the text images.
Our method achieves state-of-the-art performance in Scene Text Recognition, Removal, and Editing.

\end{abstract}    
\section{Introduction}
\label{sec:intro}

\begin{figure}[t]
  \centering
   \includegraphics[width=1.0\linewidth]{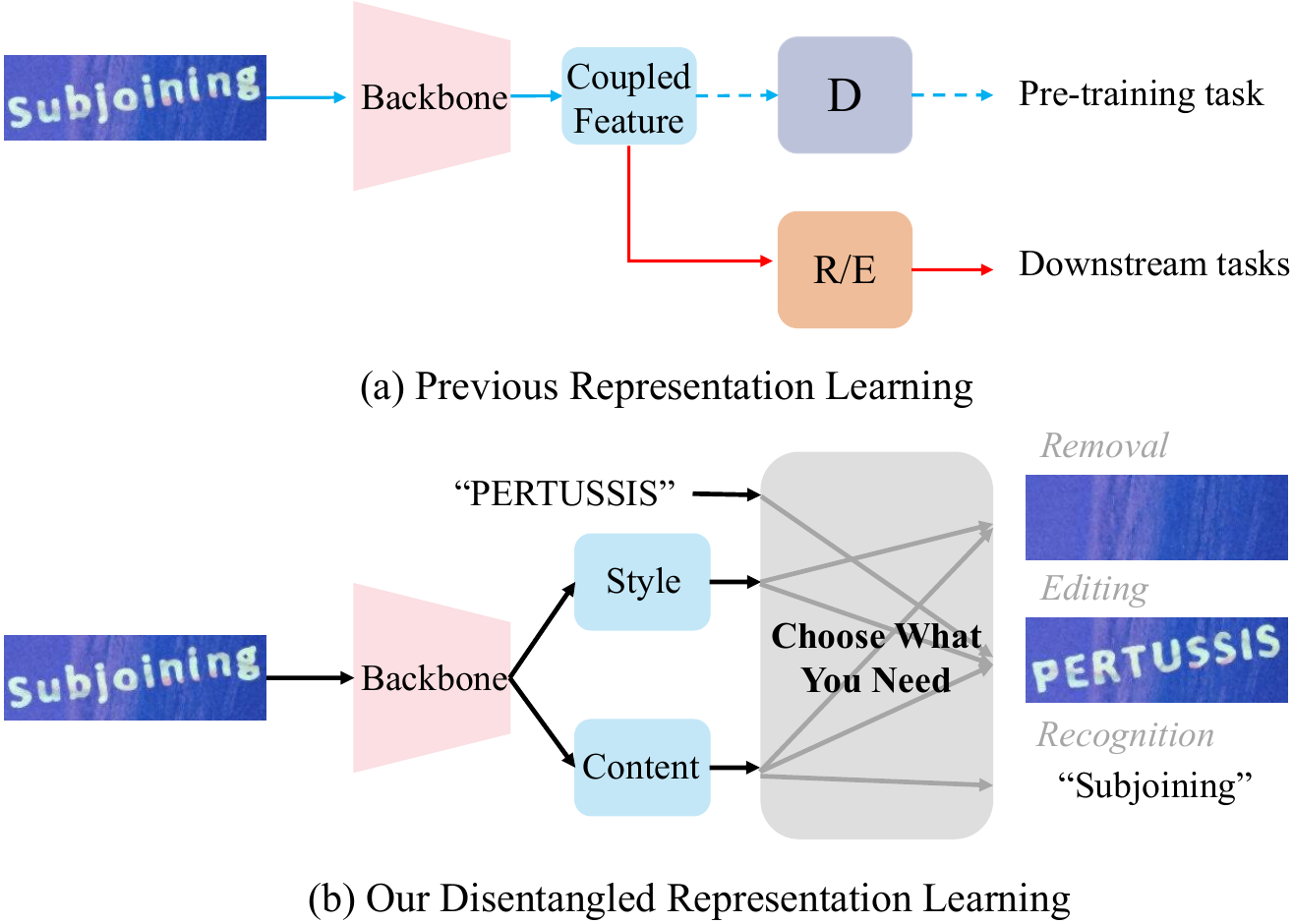}
   \caption{(a) The pipeline of previous representation learning methods that use a tightly coupled feature for all tasks. 'D' means decoder and 'R/E' represents the recognizer or eraser. (b) Our decoupled representation learning framework for multi-tasking.}
   \label{fig:intro}
\end{figure}

As an important information carrier, language is widely found in natural scenes. Scene text is a significant topic in scene understanding and perception. There have been a number of researches on scene text including Scene Text Recognition (STR), Scene Text Editing (STE), Scene Text Removal (STRM), \etc. These researches are widely used in human–computer interaction~\cite{anytext,textdiffuser2}, cross-modal understanding~\cite{ProST,momentdiff,liu2023towards,layoutlmv3}, automatic pilots, \etc.

In contemporary research, numerous studies try to leverage representation learning to enhance feature quality, thereby improving performance in downstream tasks. Within the domain of scene text, approaches employing mask image modeling (MIM) and feature contrastive learning (CL) have garnered notable success~\cite{union14m,dig}. As depicted in \cref{fig:intro} (a), these works first use a decoder to pre-train the backbone, facilitating the completion of the pretasks. Then, the pre-trained backbone is used for fine-tuning with a task-specific decoder. Although achieve impressive performance, it is apparent that these pipelines face challenges. Using the same representation for different downstream tasks including discriminative and generative is sub-optimal, limiting the generalization of the methods.

To address the above issue, we investigate distinctive properties of scene text images that set them apart from general scene images. Specifically, cropped scene text images contain a focused region with high information density alongside a diverse background. We categorize these distinctive attributes into style and content features. Style features encompass the background and text style elements (such as font, color, tilt, \etc.), while content features encompass content and texture details. In STR, the essential information is content features. Style information is considered as noise that hinders accurate recognition. Nevertheless, in STE, the process can be divided into two stages: text removal and text rendering. The text removal stage resembles an image inpainting task that reconstructs the background pixels in the original text area. This process requires content features for stroke localization and style features for background reconstruction. The text rendering stage relies on content features to generate text strokes and style features to define the font. Therefore, different downstream tasks need different information (\cref{fig:intro} (b)), and features irrelevant to a particular task may hinder task completion. The decomposition of these two types of features can be applied to various scene text tasks.

In this paper, we explore the representation learning from a novel perspective by considering the above unique properties of scene text. To decouple two types of features, we introduce a method as illustrated in Figure \ref{fig:pipeline}. To begin with, we synthesized a dataset containing pairs of images to facilitate the design of subsequent decoupled pre-training methods. The image pairs contain identical backgrounds and styles but differ in content. These pairs are simultaneously inputted into the network. Features of these image pairs are extracted using a ViT-based backbone along with a decoupling block. We directly divide the output tokens into two parts and subsequently achieve decoupling through different losses. Multi-task decoder utilizes these decoupled features to perform various tasks. To achieve the intended disentanglement of features and effectively train the model, we propose a training paradigm.
This paradigm involves exclusively using the content features for recognition while aligning the style features of the image pairs. Additionally, style features are used to reconstruct the counterpart image with a text prompt.
After pre-training, the multi-task decoder can effectively handle both generative and discriminative tasks, while also serving as a great starting for fine-tuning. Compared with previous methods, our method can accomplish both generative and discriminative tasks without the need for additional modules in the pre-training stage.

To summarize, our contributions are as follows:

\begin{itemize}
    \item We propose to decouple the features for scene text tasks, leveraging the distinctive properties of scene text images. This endeavor may encourage the research community to reconsider the distinctiveness of textual images.
    \item We propose a training paradigm for feature disentanglement in scene text images.
    \item We propose a new synthetic dataset for the training and evaluating of scene text editing.
\end{itemize}

Compared with previous methods, our DARLING achieves state-of-the-art performance in scene text recognition, editing, and removal.

\section{Related Works}
\subsection{Scene Text Tasks}
We mainly review three widely researched scene text tasks: Scene Text Recognition, Editing, and Removal.

\textbf{Scene Text Recognition (STR)} has been a significant research term in computer vision. There are three kinds of methods: CTC-based, attention-based, and segmentation-based. (1) CTC-based methods \cite{CRNN,ma2020joint,svtr} use a Connectionist Temporal Classification (CTC)~\cite{CTC} decoder to directly transform the image features into text sequences. Such an operation has a fast inference speed but a limited performance. (2) Attention-based methods \cite{aster,nrtr,cdistnet,abinet,LPV,parseq,MGP,clipstr} use a learnable query and a cross-attention operation to decode the recognition result. Such methods have high recognition accuracy but usually not as fast as CTC-based methods. (3) Segmentation-based methods \cite{liao2019scene,peng2022recognition} treat the scene text images as a combination of characters. Recently, some works \cite{guan1,guan2} use pre-processing algorithms to obtain the segmentation map which can further direct the model to get better performance. Our method uses the attention-based decoder because of its high performance and elegant consistency with the transformer structure.

\textbf{Scene Text Editing (STE)} aims to replace text in a scene image with new text while maintaining the original background and styles. Due to the great development of Generating Adversarial Networks (GAN) \cite{gan}, GAN-based scene text editing methods attract increasing research interest. SRNet \cite{srnet} first proposes to divide the editing task into three sub-process: background inpainting, text conversion and fusion, which inspires many subsequent works \cite{STEFANN,swaptext,mostel}.
Nowadays, with the advances in diffusion models, some works \cite{diffute,textdiffuser,textdiffuser2,anytext} use the diffusion process to achieve excellent editing results. However, due to the complexity of the diffusion model, these methods are less efficient. From a unified perspective, we abandon the use of diffusion and GAN, in favor of an efficient attention structure that achieves high-quality generation through decoupled features.

\begin{figure*}[t]
  \centering
   \includegraphics[width=0.9\linewidth]{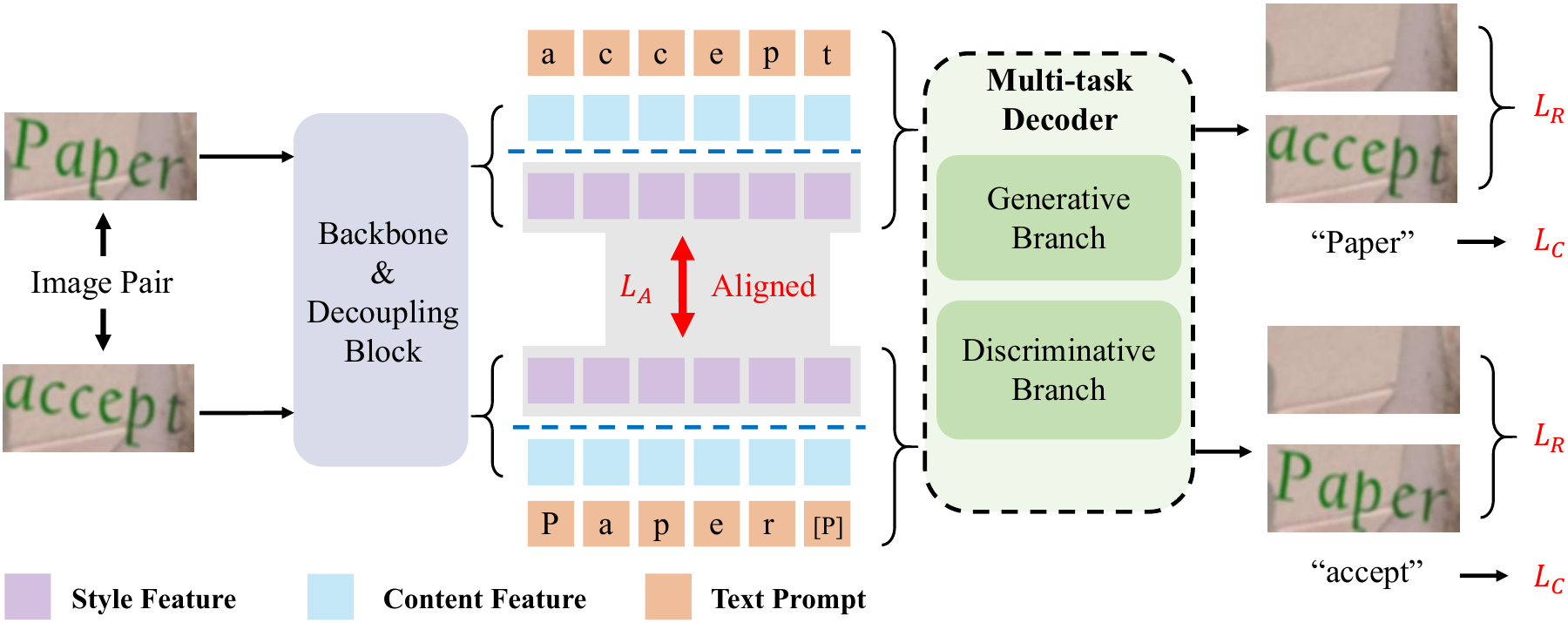}
   \caption{The pipeline and training paradigm of our DARLING. The Decoupling Block divides features from the backbone into style and content features. The multi-task decoder processes these features to perform both discriminative and generative tasks. '[p]' is the padding symbol. Image pairs with the same style but different content are input. The style features are aligned and recognition loss supervises the content features to eliminate the style from content features.}
   \label{fig:pipeline}
\end{figure*}

\textbf{Scene Text Removal (STRM)} can be seen as the first step in STE. Recently a variety of approaches \cite{mtrnet,enstext,syn,pert,fetnet,pen,ctrnet,viteraser} tried to accomplish text removal on a whole scene image. However, as texts are sparse in the scene images, text erasing directly on the scene image tends to affect non-text areas and seems uncontrollable. For uniformity, our approach uses cropped images.

\subsection{Representation Learning for Scene Text Tasks}
Several studies employ representation learning to enhance features for scene text tasks. MaskOCR \cite{maskocr} employs masked image modeling (MIM) to glean implicit knowledge from substantial unlabeled data. DiG \cite{dig} proposes the combined use of MIM and contrastive learning to bolster backbone representation for accurate text recognition. \cite{union14m} introduces a high-quality dataset for unsupervised pre-training, leading to significant improvements in STR. In the field of STRM, recent efforts \cite{pert,pen,viteraser} leverage synthetic datasets for pre-training, enhancing performance and yielding more robust features. However, these methods focus on using a tightly coupled feature to accomplish downstream tasks, which limits the generalization of the method. SimAN \cite{siman} introduces a Similarity-Aware Normalization module, implicitly decomposing features and achieving notable performance across some tasks. Despite this, the decomposition using instance normalization is not thorough. We first propose to address multiple tasks by feature disentanglement and introduce a disentangled training paradigm.

\section{Method}
Our DARLING is a pre-training method for scene text tasks including STR, STE, and STRM. In this section, we detail the pipeline of the proposed method in \cref{sec:pipeline}. Then, we propose the disentanglement training paradigm and multi-task decoder in \cref{sec:ECR} and \cref{sec:MTD}. Finally, we describe the training objective in \cref{sec:to}.

\subsection{Pipeline}\label{sec:pipeline}
The pipeline of our DARLING is shown in \cref{fig:pipeline}. Given a scene text image $\mathbf{I}$, the feature tokens $\mathbf{F}\in\mathbb{R}^{L\times D}$ are first extracted by a transformer-based backbone. Then, the features are fed into a decoupling block. We directly separate the output of the decoupling block $\mathbf{F}_D$ into two components denoted as $\mathbf{F}_D=[\mathbf{F}_S, \mathbf{F}_C]$, where $\mathbf{F}_S\in\mathbb{R}^{\frac{L}{2}\times D}$ indicates the style features and $\mathbf{F}_C\in\mathbb{R}^{\frac{L}{2}\times D}$ represents the content features. The decoupling block comprises multiple self-attention layers designed to capture long-range information essential for feature separation. Subsequently, the two kinds of features are inputted into the multi-task decoder with a text prompt indicating the desired text for rendering. This decoder is capable of handling both discriminative tasks like recognition and generative tasks such as editing and removal. Meanwhile, multi-task supervision aids in constraining features, fostering feature disentanglement, and acquiring diverse features.


\subsection{Disentangled Training Paradigm}\label{sec:ECR}
To achieve the expected disentanglement of the two feature types, we introduce a pre-training paradigm. The concept is depicted in \cref{fig:pipeline}. First, we employ a synthesis engine to create pairs of images that share identical backgrounds and fonts but differ in content. Subsequently, pairs of images are simultaneously fed into our proposed network. Due to the similarity in background and font, which we categorize as style features, we employ an alignment loss $\mathcal{L}_A$ to align the style features of image pairs. The formulation is as follows:

\begin{equation}
\mathcal{L}_A=\frac{1}{2}\sum(\mathbf{F}_{S1}-\mathbf{F}_{S2})^2,
\end{equation}
where $\mathbf{F}_{S1}$ and $\mathbf{F}_{S2}$ represents the style features of two images, respectively.

Furthermore, we exchange the content of two images by using each other's content as the text prompt $\mathbf{C}_T$. The decoder is tasked to reconstruct the other image and supervised using a reconstruction loss $\mathcal{L}_R$. In the multi-task decoder, the text recognition task only uses the content feature. Therefore, the supervision of recognition loss will eliminate the style information from content features, which further ensures the decoupling. 

\textbf{Discussion.} Through the utilization of the proposed disentangled pre-training, the model is constrained to align the common style features shared between a pair of images and use it for text editing. Meanwhile, since scene text recognition is inherently a fine-grained task that solely requires content features while treating style features as noise, we employ the recognition loss to guide $\mathbf{F}_C$ toward content features. As a result of the pre-training and the set of loss functions, the intended decoupling of the two feature types is achieved. Additionally, in contrast to previous works that solely pre-train the backbone, our method includes pre-training of the multi-task decoder, which proves beneficial for subsequent fine-tuning processes.

\subsection{Multi-task Decoder}\label{sec:MTD}

\begin{figure}[t]
  \centering
   \includegraphics[width=1.0\linewidth]{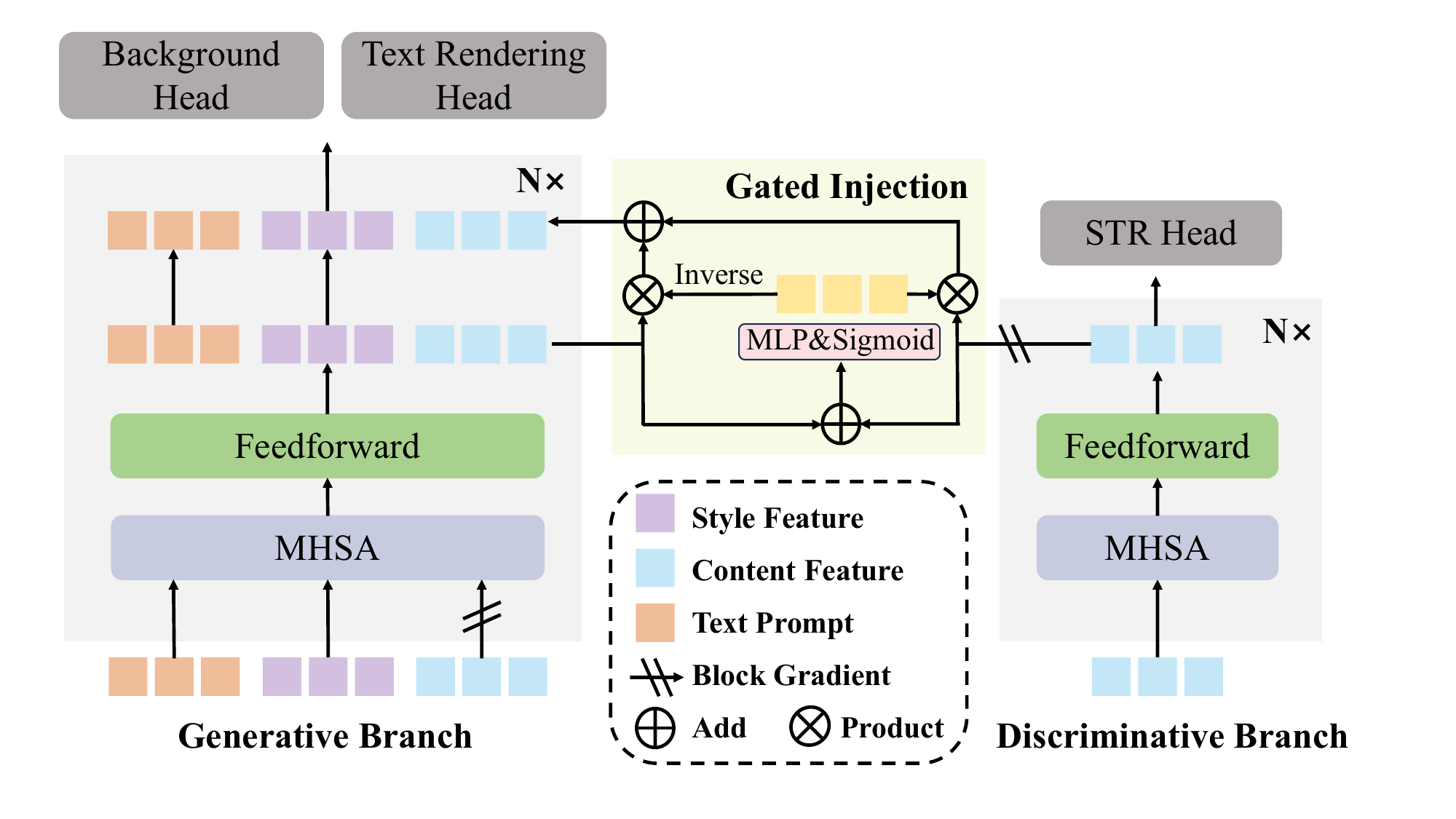}
   \caption{The structure of Multi-task Decoder. It comprises the Generative Branch (GEB) and the Discriminative Branch (DIB), each dedicated to specific tasks. Gated Injection strategy is proposed to convey fine-grained details from DIB to GEB.}
   \label{fig:MTD}
\end{figure}

The Multi-task Decoder (MTD) is designed to handle both generative tasks and discriminative tasks. As depicted in \cref{fig:MTD}, due to the distinct feature requirements of these tasks, the MTD structure comprises two branches: generative and discriminative. Additionally, we propose a gated injection strategy to integrate the information from the discriminative branch into the generative branch.

\textbf{The discriminative branch (DIB)} aims to accomplish STR task. This task demands fine-grained details but suffers from interference with style features. The input of the discriminative branch is $\mathbf{F}_C$ and the branch consists of $N$ self-attention layers to further extract the fine-grained details. Subsequently, the output features $\mathbf{F}_C^N$ are employed for text prediction by an STR head, utilizing a cross-attention operation and a linear projection. The prediction process is formalized in \cref{eq:str}, we have projection weights $\mathbf{W}_K, \mathbf{W}_V\in\mathbb{R}^{D\times D}$ and $\mathbf{W}_R\in\mathbb{R}^{D\times C}$. $\mathbf{R}\in\mathbb{R}^{T\times C}$ represents the recognition result and $\mathbf{Q}\in\mathbb{R}^{T\times D}$ is a learnable query for decoding. $T$ denotes the maximum text length, while $C$ represents the number of character classes.

\begin{equation}\label{eq:str}
\mathbf{R}=\mathbf{Q}(\mathbf{F}_C^N\mathbf{W}_K)^{\top}(\mathbf{F}_C^N\mathbf{W}_V)^{\top}\mathbf{W}_R.
\end{equation}


\textbf{The generative branch (GEB)} is designed to fulfill generative tasks such as STE and STRM. This branch is followed by two heads: the background head and the text rendering head, responsible for obtaining the background reconstruction result and the edited text image, respectively. Similar to DIB, GEB shares the same structure but receives different input. Considering the necessity for both content features in generating or removing text areas and style features in background reconstruction, we configure GEB's input as a three-part concatenation, as illustrated in \cref{eq:MTG}. $\mathbf{C}_T$ serves as the text prompt indicating the target text we intend to render on the original image.

\begin{equation}\label{eq:MTG}
input=concat(\mathbf{C}_T, \mathbf{F}_S, \mathbf{F}_C).
\end{equation}

In response to the absence of fine-grained details in $\mathbf{F}_C$, which are subsequently enhanced by DIB, we introduce a Gated Injection strategy to provide well-managed fine-grained information to GEB. To be specific, different layers of features in DIB represent information at different levels of detail, all under the supervision of recognition loss. We introduce an adaptive fusion mechanism that operates between each layer of DIB and GEB. As formalized in \cref{eq:GI}, $\odot$ represents dot production and $\mathcal{T}$ stands for the self-attention operation. The superscript indicates the layer number. $\mathbf{W}_G\in\mathbb{R}^{D\times 1}$ transforms the features into a gating weight $\mathbf{G}$, which is further used to control the fusion of $\hat{\mathbf{F}}_C$ and $\mathbf{F}_C$, where $\hat{\mathbf{F}}_C$ denotes the content features in each layer of GEB.

\begin{equation}\label{eq:GI}
\begin{aligned}
&\mathbf{G}^i=Sigmoid((\hat{\mathbf{F}}_C^i+\mathbf{F}_C^i)\mathbf{W}_G), \\
&\mathbf{F}_G^i=\hat{\mathbf{F}}_C^i\odot(1-\mathbf{G}^i)+\mathbf{F}_C^i\odot\mathbf{G}^i, \\
&[\mathbf{C}_T^i, \mathbf{F}_S^i, \hat{\mathbf{F}}_C^i]=\mathcal{T}([\mathbf{C}_T^{i-1}, \mathbf{F}_S^{i-1}, \mathbf{F}_G^{i-1}]).
\end{aligned}
\end{equation}

Finally, the background head and the text rendering head leverage a cross-attention operation to generate the residual of the background and the text expected, similar to \cref{eq:MTG}.

Overall, in our MTD, the decoupled learning approach guides the model in acquiring more discriminative features, while diverse task-guided training enhances feature diversity. Clearly, The recognition loss facilitates the decoupling of features, and the gated injection strategy combines diverse fine-grained details supervised by recognition loss to better accomplish generative tasks.

\subsection{Training Objective}\label{sec:to}
As shown in \cref{eq:loss}, the final objective function of the proposed method contains three parts: reconstruction loss $\mathcal{L}_R$, scene text recognition loss $\mathcal{L}_C$, and feature align loss $\mathcal{L}_A$.

\begin{equation}\label{eq:loss}
\mathcal{L}=\mathcal{L}_R+\lambda_C\mathcal{L}_C+\lambda_A\mathcal{L}_A.
\end{equation}

The scene text recognition loss is a cross-entropy loss which is formulated in \cref{eq:lc}, where $G_t$ is the ground truth, $T$ is the max length of the character sequence.

\begin{equation}\label{eq:lc}
\mathcal{L}_C=-\frac{1}{T}\sum_{i=0}^Tlog(P(\mathbf{R}|G_t)).
\end{equation}

\section{Experiment}
\subsection{Datasets}
\begin{figure}[t]
  \centering
   \includegraphics[width=1.0\linewidth]{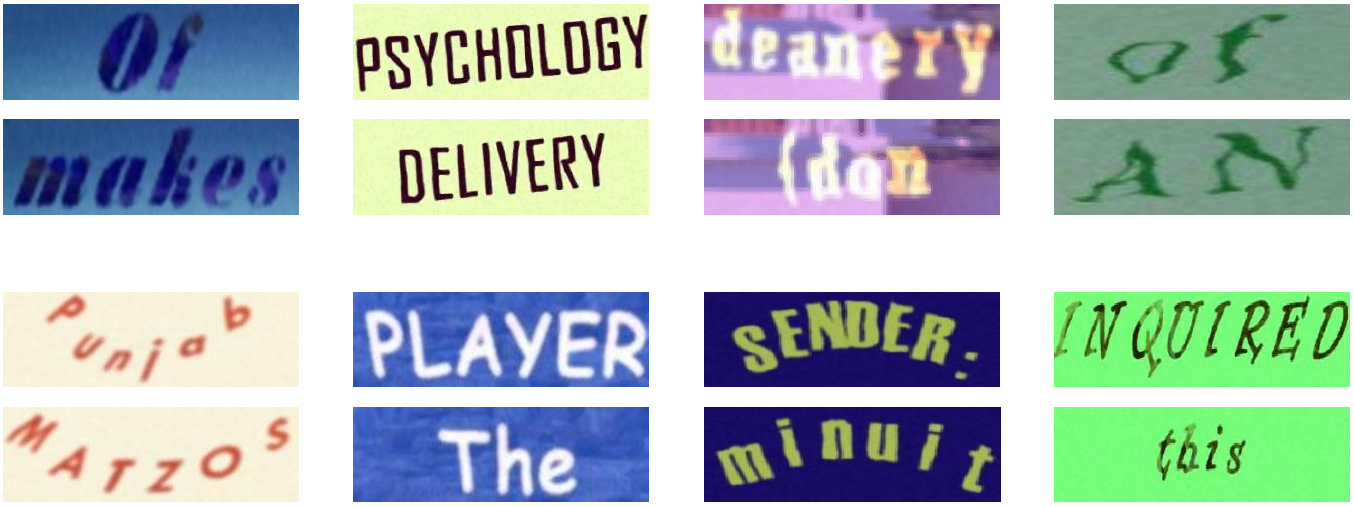}
   \caption{Some sample images from our generated datasets: TSE-4M and TSE-10k. The datasets comprise more diverse images with a variety of fonts and backgrounds, including low-quality images. TSE-10k can facilitate a more comprehensive evaluation of the model's performance.}
   \label{fig:dataset}
\end{figure}

For pre-training, we generate a dataset using the publicly available synthesis engine \cite{synthtiger}\footnote{https://github.com/clovaai/synthtiger}. We use the background images provided by SynthText \cite{ST} and the lexicon provided by MJSynth \cite{MJ1,MJ2} and SynthText \cite{ST}. The dataset contains 4M image pairs for training (TSE-4M) and 10K for evaluation (TSE-10k). Some image samples are shown in \cref{fig:dataset}. This dataset features more complex text styles and backgrounds, a broader array of fonts, and includes low-quality images affected by blurring, noise, \etc. The evaluation set serves as a more comprehensive benchmark for assessing the performance of scene text editing methods. The dataset will be made publicly available.

For the evaluation of STE, we utilize the synthetic (Tamper-Syn2k) and real (Tamper-Scene) datasets introduced by MOSTEL \cite{mostel} along with our STE-10k. For STR, we conduct fine-tuning on Union-L \cite{union14m} for fair comparison. Then, the performance is evaluated on 7 commonly used benchmarks including Union-benchmark\cite{union14m}, IIIT 5K-Words (IIIT5K)~\cite{IIIT}, ICDAR2013 (IC13)~\cite{IC13}, ICDAR2015 (IC15)~\cite{IC15}, Street View Text (SVT)~\cite{SVT}, Street View Text-Perspective (SVTP)~\cite{SVTP}, and CUTE80 (CUTE)~\cite{CUTE}. Comprehensive details regarding these datasets can be found in prior works \cite{MGP,abinet}. As for STRM, we fine-tune and evaluate the removal part of our model on SCUT-EnsText \cite{enstext}.

\subsection{Implementation Details}
In our implementation, The layer number of the decoupling block is set to 3 and the layer number of the multi-task decoder is set to 6. Following most previous methods, the image size is fixed at 128 × 32. We conduct the experiments on 4 NVIDIA 4090 GPUs with a batch size of 384. For scene text recognition, the vocabulary size C is set to 98, including 0 - 9, a - z, A-Z, special characters, [PAD] for padding symbol and [EOS] for ending symbol. $\lambda_A$ and $\lambda_C$ are both set to 0.5.

The network is pre-trained end-to-end using Adam~\cite{adam} optimizer with an initial learning rate set at 1e-4. The learning rate is then decayed to 1e-5 after seven epochs. The pre-training phase encompasses 200K iterations. For fine-tuning in the STR task, an additional 400K iterations are executed, maintaining the same learning rate schedule as in pre-training. For STRM, we fine-tuned 100K iterations.

\subsection{Comparisons with State-of-the-Arts}
Our multi-task decoder is adept at handling both discriminative (STR) and generative (STE, STRM) tasks. In this section, we conduct a series of experiments to prove the effectiveness of the MTD and our disentangled representation learning framework.

\subsubsection{Scene Text Recognition}

\begin{table*}[!t]
\caption{Comparison with state-of-the-art STR methods on Common benchmarks and Union-14M Benchmarks. \dag stands for reproducing by ourselves. \ddag means the method has a pre-training stage. All methods are trained or fine-tuned on Union-L \cite{union14m}.}\label{tab:str}
\centering
\resizebox{\textwidth}{!}{%
\begin{tabular}{l|ccccccc|cccccccc|c}
\hline
\multirow{2}{*}{Methods}           & \multicolumn{7}{c|}{Common Benchmarks}                                                                                                                                                                                                                                                                                                          & \multicolumn{8}{c|}{Union14M-Benchmark}                                                                                                                                                                                                                      & \#Params \\ \cline{2-16}
                                   & \begin{tabular}[c]{@{}c@{}}IIIT5K\\ 3000\end{tabular} & \begin{tabular}[c]{@{}c@{}}IC13\\ 1015\end{tabular} & \begin{tabular}[c]{@{}c@{}}SVT\\ 647\end{tabular} & \begin{tabular}[c]{@{}c@{}}IC15\\ 2077\end{tabular} & \begin{tabular}[c]{@{}c@{}}SVTP\\ 645\end{tabular} & \begin{tabular}[c]{@{}c@{}}CUTE\\ 288\end{tabular} & AVG           & Curve         & \begin{tabular}[c]{@{}c@{}}Multi-\\ Oriented\end{tabular} & Artistic      & \begin{tabular}[c]{@{}c@{}}Context-\\ less\end{tabular} & Salient       & \begin{tabular}[c]{@{}c@{}}Multi-\\ Words\end{tabular} & General       & AVG           & (M)      \\ \hline
CRNN~\cite{CRNN}                   & 91.8                                                  & 91.8                                                & 83.8                                              & 71.8                                                & 70.4                                               & 80.9                                               & 81.6          & 19.4          & 4.5                                                       & 34.2          & 44.0                                                    & 16.7          & 35.7                                                   & 60.4          & 30.7          & 8.3      \\
ASTER~\cite{aster}                 & 94.3                                                  & 92.6                                                & 88.9                                              & 77.7                                                & 80.5                                               & 86.5                                               & 86.7          & 38.4          & 13.0                                                      & 41.8          & 52.9                                                    & 31.9          & 49.8                                                   & 66.7          & 42.1          & 27.2     \\
NRTR~\cite{nrtr}                   & 96.2                                                  & 96.9                                                & 94.0                                              & 80.9                                                & 84.8                                               & 92.0                                               & 90.8          & 49.3          & 40.6                                                      & 54.3          & 69.6                                                    & 42.9          & 75.5                                                   & 75.2          & 58.2          & -        \\
SATRN~\cite{satrn}                 & 97.0                                                  & 97.9                                                & 95.2                                              & 87.1                                                & 91.0                                               & 96.2                                               & 93.9          & 74.8          & 64.7                                                      & 67.1          & 76.1                                                    & 72.2          & 74.1                                                   & 75.8          & 72.1          & -        \\
RobustScanner~\cite{robustscanner} & 96.8                                                  & 95.7                                                & 92.4                                              & 86.4                                                & 83.9                                               & 93.8                                               & 91.2          & 66.2          & 54.2                                                      & 61.4          & 72.7                                                    & 60.1          & 74.2                                                   & 75.7          & 66.4          & -        \\
SVTR-S~\cite{svtr}                 & 95.9                                                  & 95.5                                                & 92.4                                              & 83.9                                                & 85.7                                               & 93.1                                               & 91.1          & 72.4          & 68.2                                                      & 54.1          & 68.0                                                    & 71.4          & 67.7                                                   & 77.0          & 68.4          & 10.3     \\ \hline
SRN~\cite{SRN}                     & 95.5                                                  & 94.7                                                & 89.5                                              & 79.1                                                & 83.9                                               & 91.3                                               & 89.0          & 49.7          & 20.0                                                      & 50.7          & 61.0                                                    & 43.9          & 51.5                                                   & 62.7          & 48.5          & 55       \\
VisonLAN~\cite{visionlan}          & 96.3                                                  & 95.1                                                & 91.3                                              & 83.6                                                & 85.4                                               & 92.4                                               & 91.3          & 70.7          & 57.2                                                      & 56.7          & 63.8                                                    & 67.6          & 47.3                                                   & 74.2          & 62.5          & 33       \\
PARSeq~\cite{parseq}\dag           & 98.0                                                  & 96.8                                                & 95.2                                              & 85.2                                                & 90.5                                               & 96.5                                               & 93.5          & 79.8          & {\ul 79.2}                                                & 67.4          & 77.4                                                    & 77.0          & 76.9                                                   & 80.6          & 76.9          & 23.8     \\
ABINet~\cite{abinet}               & 97.2                                                  & 97.2                                                & 95.7                                              & 87.6                                                & 92.1                                               & 94.4                                               & 94.0          & 75.0          & 61.5                                                      & 65.3          & 71.1                                                    & 72.9          & 59.1                                                   & 79.4          & 69.2          & 37       \\
MAERec-S~\cite{union14m}\ddag      & 98.0                                                  & 97.6                                                & 96.8                                              & 87.1                                                & {\ul 93.2}                                         & \textbf{97.9}                                      & {\ul 95.1}    & {\ul 81.4}    & 71.4                                                      & 72.0          & \textbf{82.0}                                           & {\ul 78.5}    & \textbf{82.4}                                          & {\ul 82.5}    & {\ul 78.6}    & 35.8     \\
CCD~\cite{guan2}\dag\ddag          & {\ul 98.3}                                            & 97.6                                                & {\ul 97.1}                                        & {\ul 88.3}                                          & 92.3                                               & {\ul 97.2}                                         & {\ul 95.1}    & 79.1          & 76.8                                                      & {\ul 72.2}    & 80.0                                                    & 78.0          & {\ul 80.2}                                             & 81.5          & 78.3          & 36       \\ \hline
Baseline                           & 98.1                                                  & {\ul 98.2}                                          & 96.9                                              & 87.2                                                & 91.9                                               & 96.2                                               & 94.8          & 79.4          & 71.1                                                      & 68.4          & 77.5                                                    & 75.9          & 77.5                                                   & 81.0          & 75.8          & 18.7     \\
DARLING (Ours)\ddag                & \textbf{98.5}                                         & \textbf{98.7}                                       & \textbf{97.8}                                     & \textbf{88.5}                                       & \textbf{93.3}                                      & 96.5                                               & \textbf{95.6} & \textbf{85.8} & \textbf{79.7}                                             & \textbf{72.3} & {\ul 80.4}                                              & \textbf{81.1} & 78.7                                                   & \textbf{83.3} & \textbf{80.2} & 18.7     \\ \hline
\end{tabular}%
}
\end{table*}

\begin{table}[]
\caption{Comparison with state-of-the-art STR methods on some more complicated datasets including occlusion and wordart.}\label{tab:strchallenge}
\centering
\resizebox{\columnwidth}{!}{%
\begin{tabular}{l|ccc|c}
\hline
Methods                            & WOST          & HOST          & WordArt       & \multicolumn{1}{l}{\# Params(M)} \\ \hline
CRNN~\cite{CRNN}                   & -             & -             & 47.5          & 8.3                              \\
ASTER~\cite{aster}                 & -             & -             & 57.9          & 27.2                             \\
RobustScanner~\cite{robustscanner} & -             & -             & 61.3          & -                                \\
VisonLAN~\cite{visionlan}          & 70.8          & 49.8          & 69.1          & 33                               \\
ABINet~\cite{abinet}               & 75.3          & 57.9          & 67.4          & 37                               \\
PARSeq~\cite{parseq}               & 73.6          & 55.4          & 79.2          & 23.8                             \\
MGP-Small~\cite{MGP}               & 76.0          & 62.8          & 69.0          & 52.6                             \\
SVTR-S~\cite{svtr}                 & 74.6          & 58.9          & 65.9          & 10.3                             \\
CCD~\cite{guan2}\dag\ddag          & 80.6          & 67.3          & 79.3          & 36                               \\ \hline
Baseline                           & 78.3          & 62.8          & 78.7          & 18.7                             \\
DARLING (Ours)\ddag                & \textbf{82.5} & \textbf{70.8} & \textbf{81.2} & 18.7                             \\ \hline
\end{tabular}%
}
\end{table}

For the task of scene text recognition, we evaluated our approach on six widely used benchmarks (common benchmarks) as well as the more diverse Union14M-Benchmark \cite{union14m}. The results are shown in \cref{tab:str}, the Baseline is the result of our model trained on STE-4M and Union-L without disentangled pre-train. Compared with it, our pre-trained model obtains a remarkable performance enhancement across all datasets. Compared with other methods, our approach surpasses the 0.5\% average accuracy of the state-of-the-art model while employing fewer parameters. This outcome substantiates that our approach attains a high-quality representation adept at effectively handling diverse text images in real-world scenes. The Union14M-Benchmark encompasses a variety of challenging images, such as curved, multi-oriented, artistic, and salient text. Our significant performance improvement is evident across these datasets. Notably, our approach outperforms the SOTA performance by 4.4\%, 2.6\%, and 0.5\% on curved, salient, and multi-oriented datasets, respectively. Our approach demonstrates a slightly diminished performance on multi-word images which contain several words within a single image. This limitation can be mitigated by employing a robust text detector.

Compared with other pre-training methods, our approach still exhibits high performance. Note that we just pre-train our model on 4M synthetic image pairs we proposed. In contrast, other pre-training methods rely on approximately 10M \textbf{real} images for pre-training. This dataset is considerably more difficult to acquire than our synthetic data.

Furthermore, we conduct experiments on more challenging datasets in \cref{tab:strchallenge}. These datasets encompass scenes with occlusions (WOST, HOST \cite{visionlan}) and art characters (WortArt \cite{wordart}). A more discriminative representation is required in these scenarios, and our approach achieves the best performance (1.9\%, 3.5\%, 1.9\% performance gain) compared to previous methods. Despite being a language-independent framework, our approach yields high-quality features that offer dependable information in scenarios involving occlusion and art words.

\subsubsection{Scene Text Editing}

\begin{table*}[]
\caption{Comparison with state-of-the-art STE methods on synthetic and real datasets. \dag stands for the methods we reproduce. ClassAcc is the metric we proposed to evaluate the realism of the generated images. The SSIM, SeqAcc, and ClassAcc are presented in percent (\%).}\label{tab:ste}
\centering
\resizebox{\textwidth}{!}{%
\begin{tabular}{l|ccccc|cc|ccccc}
\hline
\multirow{2}{*}{Methods}     & \multicolumn{5}{c|}{Tamper-Syn2k}                                                   & \multicolumn{2}{c|}{Tamper-Scene}   & \multicolumn{5}{c}{STE-10k}                                                         \\ \cline{2-13} 
                             & MSE$\downarrow$   & PSNR$\uparrow$   & SSIM$\uparrow$   & FID$\downarrow$  & SeqAcc$\uparrow$ & SeqAcc$\uparrow$ & ClassAcc$\downarrow$ & MSE$\downarrow$   & PSNR$\uparrow$   & SSIM$\uparrow$   & FID$\downarrow$  & SeqAcc$\uparrow$ \\ \hline
pix2pix\dag                  & 0.1450          & 9.18           & 34.15          & 127.21         & 6.1            & 13.26          & 94.74                  & 0.1291          & 9.78           & 31.69          & 132.52         & 12.2           \\
SRNet~\cite{srnet}\dag       & 0.0216          & 18.66          & 49.97          & 64.37          & -              & 30.36          & -                  & -               & -              & -              & -              & -              \\
SwapText~\cite{swaptext}\dag & 0.0194          & 19.43          & 52.43          & 53.23          & -              & 54.83          & -                  & -               & -              & -              & -              & -              \\
MOSTEL~\cite{mostel}         & 0.0135          & 20.27          & 56.94          & \textbf{33.79} & 35.9           & 66.54          & 69.81              & 0.0169          & 19.24          & 43.56          & \textbf{18.46} & 50.71          \\
DARLING (Ours)               & \textbf{0.0120} & \textbf{20.80} & \textbf{60.07} & 44.48          & \textbf{38.3}  & \textbf{70.85} & \textbf{66.46}     & \textbf{0.0100} & \textbf{21.77} & \textbf{59.95} & 37.75          & \textbf{61.00} \\ \hline
\end{tabular}%
}
\end{table*}

\begin{figure*}[t]
  \centering
   \includegraphics[width=1.0\linewidth]{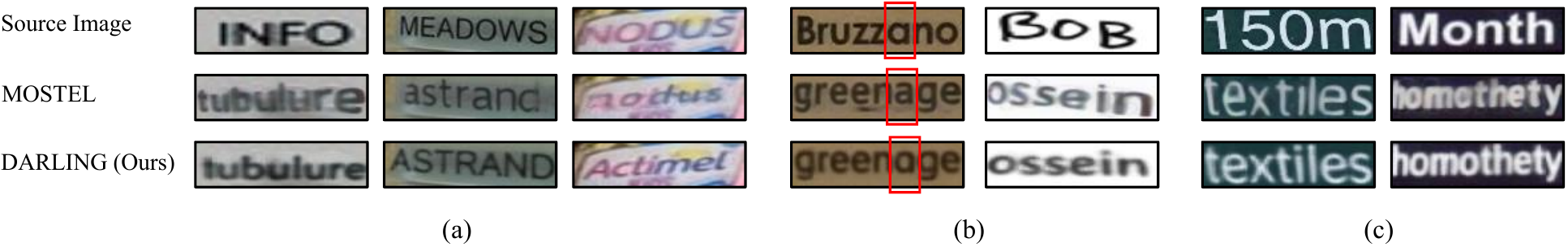}
   \caption{Qualitative examples of text editing in real scenes. (a) Comparison of the generation quality. (b) Comparison of the ability to maintain style. (c) Comparison of the realism when the generated images are clear and readable.}
   \label{fig:steoverall}
\end{figure*}

\begin{figure}[t]
  \centering
   \includegraphics[width=0.9\linewidth]{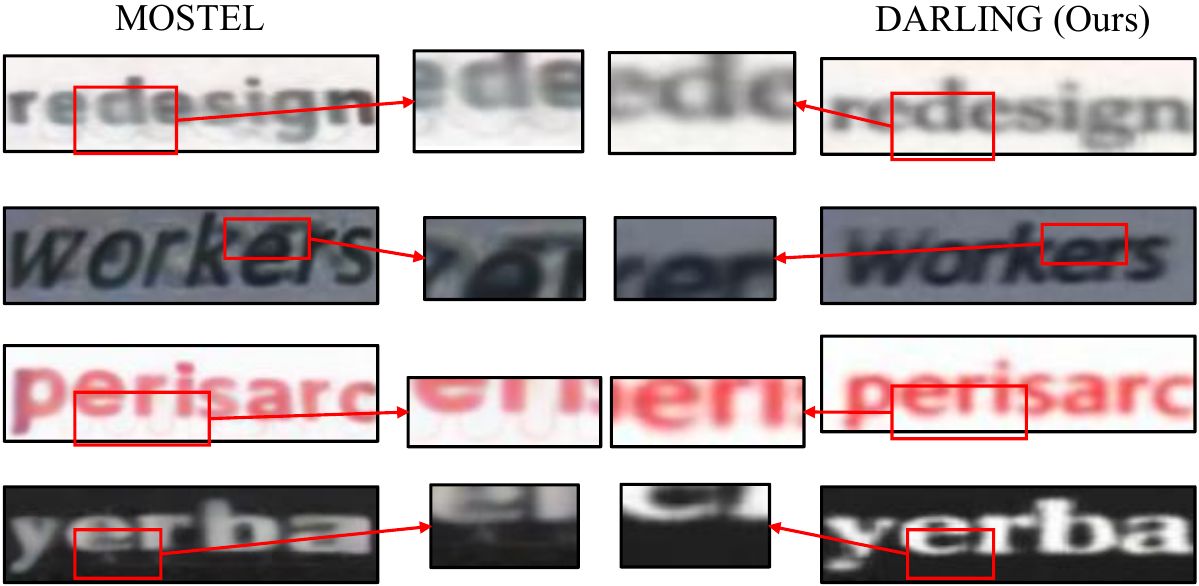}
   \caption{Comparison of generating details in real scenes. Details like artifacts, textures, and sharp edges lead to the fake appearance.}
   \label{fig:stedetail}
\end{figure}

Our method uses a disentangled training paradigm, allowing for the direct acquisition of a scene text editing network without the need for fine-tuning. We assess its performance on the datasets introduced by MOSTEL \cite{mostel} and our STE-10k. To comprehensively evaluate the edited images of our method on synthetic datasets, we adopt the following commonly used metrics: 1) MSE, the $L2$ distances; 2) PSNR, the ratio of peak signal to noise; 3) SSIM, the mean structural similarity; 4) FID, the distances between features extracted by Inception V3. In experiments with real-scene datasets, the absence of ground truth poses a challenge. We employ a metric named SeqAcc, which measures recognition accuracy using a widely utilized OCR engine\footnote{https://github.com/PaddlePaddle/PaddleOCR}. Furthermore, we found in our experiments that the above metrics being good does not mean that the edited images are of high quality. In order to better assess the authenticity of edited images, we propose a new metric called ClassAcc. Specifically, we employ a simple convolutional network trained on the original and edited images to distinguish the authenticity of the images. The accuracy of this network's classification acts as the metric for edit quality, where lower accuracy represents more authentic generated images, thus indicating superior editing quality. For fair comparison, we resize the images to $128\times32$. The results are shown in \cref{tab:ste}.

It's noteworthy that other methods often necessitate more supervision and pre-training. For instance, SRNet demands supervision for text masks, text skeleton images, and standard text images, along with the use of a discriminator for adversarial training. MOSTEL, on the other hand, requires supervision for text masks and pre-training for both text removal and text recognition models. In contrast, our approach solely relies on image pairs and does not require a discriminator or additional pre-training.

Compared with other approaches, our method achieves significant performance on both synthetic and real datasets. On synthetic datasets, we have the best performance except FID score. This is due to the constraints in the synthesizer's synthesis quality, causing the ground truth image to not always be the most realistic and unique. Due to the limits of synthetic images, real-world scenarios can better reflect the generative capability. Our model attains 10.94\% SeqAcc improvement on the real dataset (Tamper-Scene), showcasing the readability of our outputs. As shown in \cref{fig:steoverall} (a), the generated images exhibit more accurate text pixels. Additionally, our method effectively preserves the style of the source image, as demonstrated in \cref{fig:steoverall} (b). Furthermore, we argue that achieving readability and correct stylization in generated images is insufficient for measuring generation authenticity. As shown in \cref{fig:steoverall} (c), in certain instances, the generated images are easily recognizable as fake. Hence, we introduce the ClassAcc metric, and our method surpasses the state-of-the-art by 3\%. We further magnify the local details of the generated images in \cref{fig:stedetail}. In these instances, details like artifacts, textures, and sharp edges lead to fake appearance, but our results remain high quality.

\subsubsection{Scene Text Removal}
\begin{figure}[t]
  \centering
   \includegraphics[width=1.0\linewidth]{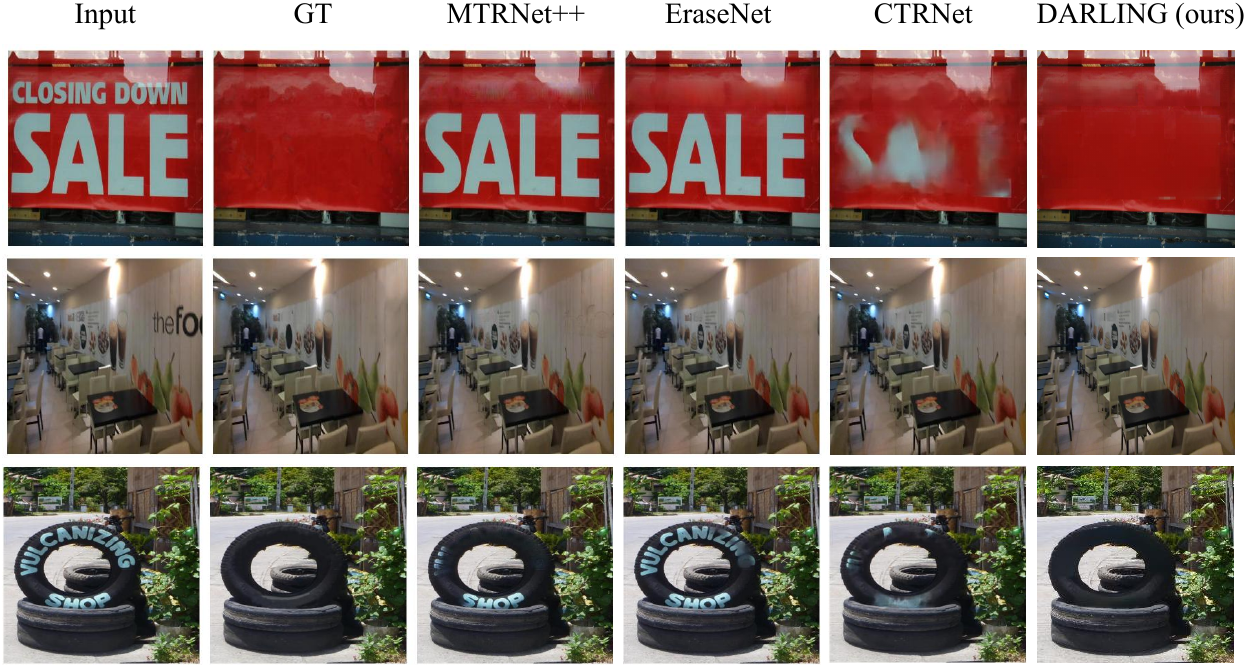}
   \caption{Some qualitative examples in STRM task.}
   \label{fig:strm}
\end{figure}

Our model incorporates a background head in the multi-task decoder, enabling it to perform the scene text removal task. We feed the image into the model with the text prompt $\mathbf{C}_T$ set as '[B][E][P][P]...' to complete the STRM task, where '[B]', '[E]', and '[P]' represent the beginning, end, and padding symbols, respectively. The evaluation was conducted on the widely used real scene dataset SCUT-EnsText \cite{enstext}. The results are detailed in \cref{tab:strm}. In comparison to our baseline without disentangled pre-training, our pre-training approach demonstrates a noteworthy improvement. This underscores the advantageous impact of disentangled pre-training on representation learning. When compared to state-of-the-art methods, our approach achieves the best performance across all metrics. Although our method of erasing a cropped image is naturally better than the previous method of erasing over the entire image, we can still see the superior erasing power of our method from the table comparison. Moreover, some qualitative examples are shown in \cref{fig:strm}, our method can obtain a more effective removal outcome while keeping the unrelated areas unaffected.

\subsection{Ablation Study}
\begin{figure}[t]
  \centering
   \includegraphics[width=1.0\linewidth]{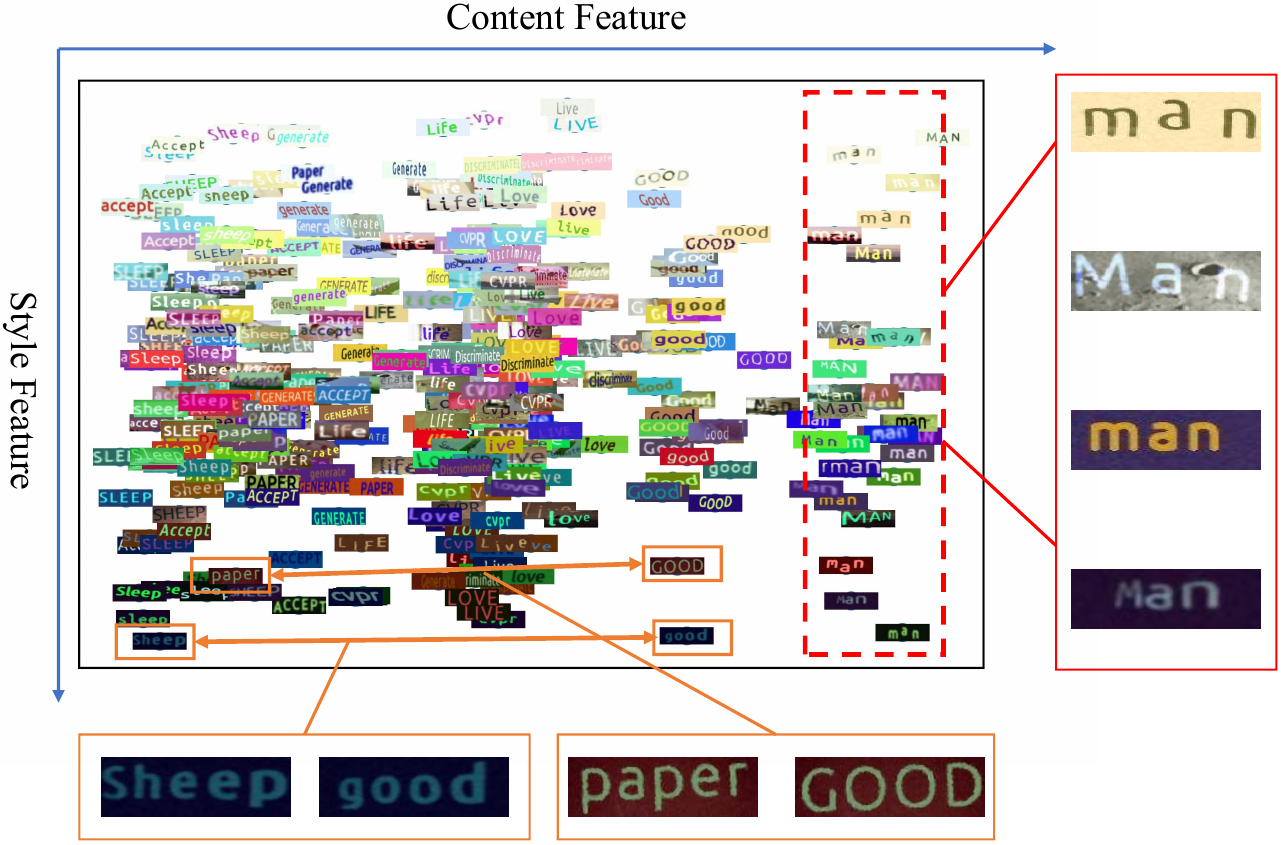}
   \caption{Feature visualization about the disentangled capabilities of our pre-trained model.}
   \label{fig:deouple}
\end{figure}

\textbf{The disentanglement of our model:} We propose a disentangled representation learning framework for scene text tasks. Leveraging the disentangled training paradigm, we align style features across style-consistent image pairs. The recognition loss then guides content features to eliminate style information, achieving effective decoupling. The superior performance across various tasks substantiates the effectiveness of our disentangled representation learning paradigm. We further visualize the features in a scatterplot with the help of PCA algorithm to validate the decoupling capability of our model. Some simple images are depicted in \cref{fig:deouple}. Along the axis of the style feature, different positions roughly correspond to different backgrounds and fonts, and along the axis of the content features, different positions roughly correspond to different contents. This visualization affirms that the representation of two types of features differs significantly. When handling diverse tasks, we selectively employ different features to achieve notable performance improvements.

\begin{table}[]
\caption{Comparison with state-of-the-art STRM methods on the real scene dataset SCUT-EnsText \cite{enstext}. \ddag means the methods with a pre-training stage. The SSIM is presented in percent (\%).}\label{tab:strm}
\centering
\resizebox{\columnwidth}{!}{%
\begin{tabular}{l|cccc|c}
\hline
Methods                              & MSE$\downarrow$   & PSNR$\uparrow$   & SSIM$\uparrow$   & FID$\downarrow$  & \# Params \\ \hline
pix2pix                              & 0.0037          & 26.70          & 88.56          & 46.88          & 54.42     \\
EnsNet~\cite{syn}                    & 0.0024          & 29.54          & 92.74          & 32.71          & 12.40     \\
MTRNet++~\cite{mtrnet}               & 0.0028          & 29.63          & 93.71          & 35.68          & 18.67     \\
EraseNet~\cite{enstext}              & 0.0015          & 32.30          & 95.42          & 19.27          & 17.82     \\
PERT~\cite{pert}                     & 0.0014          & 33.25          & 96.95          & -              & 14.00     \\
CTRNet~\cite{ctrnet}                 & 0.0009          & 35.20          & 97.36          & 13.99          & 159.81    \\
FETNet~\cite{fetnet}                 & 0.0013          & 34.53          & 97.01          & -              & 8.53      \\
PEN~\cite{pen}\ddag                  & \textbf{0.0005} & 35.72          & 96.68          & -              & -         \\
ViTEraser-Tiny~\cite{viteraser}\ddag & \textbf{0.0005} & 36.80          & 97.55          & 10.79          & 65.39     \\ \hline
Baseline                             & 0.0011          & 33.11          & 95.90          & 11.03          & 23.70     \\
DARLING (Ours)                       & \textbf{0.0005} & \textbf{38.85} & \textbf{98.25} & \textbf{10.11} & 23.70     \\ \hline
\end{tabular}%
}
\end{table}

\begin{figure}
\begin{minipage}[b]{0.43\columnwidth}
\centering
\resizebox{\columnwidth}{!}{%
\includegraphics{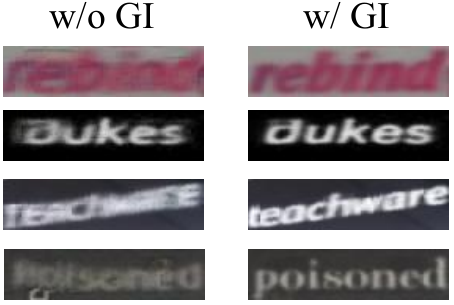}
}
\captionof{figure}{Some samples about the gated injection strategy. "w/o GI" is without gated injection. "w/ GI" means using gated injection.}\label{fig:gi}
\end{minipage}%
\hfill
\begin{minipage}[b]{0.53\columnwidth}
\centering
\resizebox{\columnwidth}{!}{
    \begin{tabular}{l|ccc}
    \hline
    \multirow{2}{*}{Methods} & \multicolumn{3}{c}{\begin{tabular}[c]{@{}c@{}}Tamper-Syn2k\\ STE-10k\end{tabular}} \\ \cline{2-4} 
                             & MSE$\downarrow$              & PSNR$\uparrow$              & SSIM$\uparrow$              \\ \hline
    \multirow{2}{*}{w/o GI}  & 0.0136                     & 19.56                     & 54.02                     \\
                             & 0.0148                     & 20.46                     & 57.69                     \\ \hline
    \multirow{2}{*}{w/ GI}   & \textbf{0.0120}            & \textbf{20.80}            & \textbf{60.07}            \\
                             & \textbf{0.0100}            & \textbf{21.77}            & \textbf{59.95}            \\ \hline
    \end{tabular}%
}
\captionof{table}{Ablation study of gated injection strategy. "w/o GI" denotes the absence of gated injection, while "w/ GI" means the utilization of gated injection.}\label{tab:gi}
\end{minipage}
\end{figure}

\textbf{Effectiveness of Gated Injection:}
The proposed gated injection can provide well-managed fine-grained details of different levels to assist in the generation of text pixels. To substantiate this, we conducted an ablation study (see Tab. \ref{tab:gi}) and presented examples in Fig. \ref{fig:gi}. It is evident that the inclusion of the gated injection strategy significantly enhances text editing performance, resulting in a clearer and more realistic generation of text details.

\section{Limitation}
Compared to other self-supervised works using real-world data, our method, utilizing synthetic data, offers certain advantages. Firstly, synthetic data is more easily accessible. Secondly, unlabeled real data cannot be employed to pre-train the decoder, a distinctive feature of our method. However, our approach has its limitations. Synthetic data exhibits a domain gap with real data, which may impact performance. Additionally, a substantial amount of unlabeled real data already exists. The question arises of how to integrate them into our method or directly design self-supervised decoupled representation learning.

\section{Conclusion}
we explore the distinctions between scene text and general scene images, proposing to decouple the two distinctive features (style and content) within scene text images. Employing our disentangled representation learning framework, the model acquires more discriminative features. When addressing various downstream tasks, distinct features are utilized. Specifically, for STR, only content features are employed to eliminate style interference. For generative tasks such as STE and STRM, style features coupled with well-managed content features are utilized to generate more realistic images. Our approach achieves state-of-the-art performance in STR, STE, and STRM. We believe this work offers valuable insights into differentiating scene text images from general images, fostering inspiration for future research in the field of scene text.

\noindent\textbf{Acknowledgments:} This work is supported by the National Key Research and Development Program of China (2022YFB3104700), the National Nature Science Foundation of China (U23B2028, 62121002, 62102384).

{
    \small
    \bibliographystyle{ieeenat_fullname}
    \bibliography{main}
}


\end{document}